%% file: acl_latex.tex
\pdfoutput=1
\documentclass[11pt]{article}
\usepackage{acl}
\usepackage{times}
\usepackage{latexsym}
\usepackage[T1]{fontenc}
\usepackage[utf8]{inputenc}
\usepackage{microtype}
\usepackage{amsmath}
\usepackage{braket}
\usepackage{subfigure}
\usepackage[export]{adjustbox}
\usepackage{url}
\usepackage{bm}
\usepackage{tikz}
\usepackage[ruled,vlined,algo2e]{algorithm2e}
\usepackage{algorithm}
\usepackage{algorithmic}
\usepackage{setspace}
\usepackage{hyperref}
\usepackage{enumitem}
\usepackage{makecell}
\let\Algorithm\algorithm
\renewcommand\algorithm[1][]{\Algorithm[#1]\setstretch{1}}

\usepackage{xcolor}
\usepackage[ruled,vlined,algo2e]{algorithm2e}

\SetCommentSty{mycommfont}
\AtBeginDocument{
  \let\mathbb\relax
  \DeclareMathAlphabet{\mathbb}{U}{msb}{m}{n}
}

\title{DIGAT: Modeling News Recommendation with Dual-Graph Interaction}

\author{
  Zhiming Mao$^{1,2}$, Jian Li$^{3}$, Hongru Wang$^{1,2}$, Xingshan Zeng$^4$, Kam-Fai Wong$^{1,2}$ \\
  $^1$The Chinese University of Hong Kong, Hong Kong, China \\
  $^2$MoE Key Laboratory of High Confidence Software Technologies, China \\
  $^3$Tencent, Shenzhen, China \\
  $^{1,2}$\texttt{\{zmmao,hrwang,kfwong\}@se.cuhk.edu.hk} \\
  $^3$\texttt{lijianjack@gmail.com} \quad
  $^4$\texttt{zxshamson@gmail.com} \\
}
\date{}

\begin{document}
\maketitle
\begin{abstract}
News recommendation~(NR) is essential for online news services. Existing NR methods typically adopt a news-user representation learning framework, facing two potential limitations. First, in news encoder, single candidate news encoding suffers from an \textit{insufficient semantic information} problem. Second, existing graph-based NR methods are promising but lack effective news-user feature interaction, rendering the graph-based recommendation suboptimal. To overcome these limitations, we propose dual-interactive graph attention networks~(\textit{DIGAT}) consisting of news- and user-graph channels. In the news-graph channel, we enrich the semantics of single candidate news by incorporating the semantically relevant news information with a semantic-augmented graph~(SAG). In the user-graph channel, multi-level user interests are represented in a news-topic graph. Most notably, we design a dual-graph interaction process to perform effective feature interaction between the news and user graphs, which facilitates accurate news-user representation matching. Experiment results on the benchmark dataset \textit{MIND} show that \textit{DIGAT} outperforms existing news recommendation methods\footnote{Our code is available at \href{https://github.com/Veason-silverbullet/DIGAT}{https://github.com/Veason-silverbullet/DIGAT}.}. Further ablation studies and analyses validate the effectiveness of (\romannumeral 1) semantic-augmented news graph modeling and (\romannumeral 2) dual-graph interaction.
\end{abstract}

\section{Introduction}\label{introduction}
News recommendation is an important technique to provide people with the news which satisfies their personalized reading interests~\citep{DAE_RNN, MIND}. Effective news recommender systems require both accurate textual modeling on news content~\citep{DKN, NRMS, FIM} and personal-interest modeling on user behavior~\citep{GNUD, HieRec}. Hence, most news recommendation methods~\citep{LSTUR, NAML, NPA, topicRec, NRMS, GERL, pp-Rec, HieRec} adopt a news-user representation learning framework to learn discriminative news and user representations, as illustrated in Figure~\ref{fig:news-recommendation-framework}.

\input{figs/news-recommendation-framework}
Though promising, there are still two potential limitations in the existing news recommendation framework. First, in news encoder, single candidate news encoding suffers from an \textit{insufficient semantic information} problem. Unlike \textit{long-term} items in common recommendation~(e.g., E-commerce product recommendation), the candidate news items are \textit{short-term} and suffer from the \textit{cold start} problem. In the real-world setting, news recommender systems usually handle the latest news, where existing user-click interactions are always not available\footnote{From the viewpoint of experimental dataset, most candidate news in test data does not appear in training user history.}. Hence, it is intractable to use existing user-click records to enrich the information of candidate news. On the other hand, compared to abundant historical clicked news in user encoder, the single candidate news may not contain sufficient semantic information for accurate news-user representation matching in the click prediction stage. Prior studies \citep{NAML, topicRec, HieRec} pointed out that users were usually interested in specific news topics~(e.g., \textit{Sports}). Empirically, the text of single candidate news does not contain enough syntactic and semantic information to accurately represent a genre of news topic and match user interests.

Second, previous studies generally follow two research directions to model user history, i.e., sequence and graph modeling. Formulating user history as a sequence of user's clicked news is a more prevalent direction, based on which time-sequential models~\citep{DAE_RNN, LSTUR, pp-Rec} and attentive models~\citep{DAN, NAML, NPA, NRMS, KIM, HieRec} are proposed. Besides, graph modeling is proved effective for recommender systems~\citep{graph-based-rec}. \citet{GERL} and \citet{GNUD} formulate news and users jointly in a bipartite graph to model news-user interaction. However, since most candidate news in test data has no existing interaction with users~(i.e., \textit{cold-news}), the isolated \textit{cold-news} nodes cause this bipartite graph modeling degenerate. Recent works formulate user history as heterogeneous graphs and employ advanced graph learning methods to extract the user-graph representations~\citep{GNewsRec, CNE-SUE, HGP}. These works focus on how to extract fine-grained representations from the user-graph side but neglect necessary feature interaction between the candidate news and user-graphs.

In this work, we propose \textbf{D}ual-\textbf{I}nteractive \textbf{G}raph \textbf{AT}tention networks~(\textit{DIGAT}) to address the aforementioned limitations. \textit{DIGAT} consists of news- and user-graph channels to encode the candidate news and user history, respectively. In the news-graph channel, we introduce semantic-augmented graph~(SAG) modeling to enrich the semantic representation of the single candidate news. In SAG, the original candidate news is regarded as the root node, while the semantic-relevant news documents are represented as the extended nodes to augment the semantics of the candidate news. We integrate the local and global contexts of SAG as the semantic-augmented candidate news representations.

In the user-graph channel, motivated by \citet{CNE-SUE} and \citet{HGP}, we model user history with a news-topic graph to represent multi-levels of user interests. Most notably, we design a dual-graph interaction process to learn news- and user-graph representations with effective feature interaction. Different from the individual graph attention network~\citep{GAT}, \textit{DIGAT} updates news and user graph embeddings with the interactive attention mechanism. Particularly, in each layer of the dual-graph, the user~(news) graph context is incorporated into its dual news~(user) node embedding learning iteratively.

Extensive experiments on the benchmark dataset \textit{MIND}~\citep{MIND} show that \textit{DIGAT} significantly outperforms the existing news recommendation methods. Further ablation studies and analyses confirm that semantic-augmented news graph modeling and dual-graph interaction can substantially improve news recommendation performance.

\section{Related Work}
Personalized news recommendation is important to online news services~\citep{DAE_RNN, FedRec}. Existing neural news recommendation methods typically aim to learn informative news and user representations~\citep{DKN, DAN, LSTUR, NAML, NPA, NRMS, KRED, FIM, KIM, pp-Rec, HieRec, HGP, MINER}. For example, \citet{LSTUR} used a CNN network to extract textual representations from news titles and used a GRU network to learn short-term user interests combined with long-term user embeddings. The matching probabilities between candidate news and users are computed over the learned news and user representations. \citet{NRMS} utilized multihead self-attention networks to learn informative news and user representations from news titles and user clicked history. These methods regarded the single candidate news as the input to news encoder, which may not contain sufficient semantics to represent a user-interested news topic. Different from these methods, we encode the candidate news with semantic-augmented graphs to enrich its semantic representations. More recently, graph-based methods were proposed for news recommendation~\citep{GERL, GNewsRec, GNUD, CNE-SUE, HGP}. For example, \citet{HGP} proposed a heterogeneous graph pooling method to learn fine-grained user representations. However, feature interaction between candidate news and users is inadequate or neglected in these methods. In contrast, our approach models effective feature interaction between news and user graphs for accurate news-user representation matching.

\section{Approach}
\textbf{Problem Formulation.} Denote the clicked-news history of a user $u$ as $H_{u}=[n_{1},n_{2},...,n_{|H|}]$, containing $|H|$ clicked news items. For the news $n$, its textual content consists of a sequence of $|T| $ words as $T_n=[w_{1},w_{2},...,w_{|T|}]$. Based on $H_{u}$ and $T_n$, the goal of news recommendation is to predict the score $\hat{s}_{n,u}$, which indicates the probability of the user $u$ clicking the candidate news $n_{can}$. The recommendation result is generated by ranking the user-click scores of multiple candidate news items.

\subsection{News Semantic Representation}\label{news_encoder}
We introduce how to extract semantic representation from news content text $T_{n}=[w_{1} ,w_{2},..., w_{|T|}]$. Our news encoder first maps the news word tokens into word embeddings $E_{n}=[e_{1}, e_{2},..., e_{|T|}]$. Then, we use the multihead self-attention network $\textrm{MSA}(\mathrm{Q},\mathrm{K},\mathrm{V})$ of Transformer encoder~\citep{transformer} to learn the contextual representations $H_{n}\in\mathbb{R}^{|T|\times{d}}$ (where $d$ is the feature dimension). Finally, we employ an attention network $f_{att}(\cdot)$ to aggregate the news semantic representation $h\in\mathbb{R}^{d}_{\mathlarger{\cdot}}$
\begin{equation*}\label{eq1_}
   H_{n}=\textrm{MSA}(\mathrm{Q}=E_{n}, \mathrm{K}=E_{n}, \mathrm{V}=E_{n})
\end{equation*}
\vskip -3mm
\begin{equation}\label{eq1}
    h = f_{att}\big(\mathrm{ReLU}(H_{n})\big)
\end{equation}
The attentive aggregation function $f_{att}(\cdot)$ is implemented by a feed-forward network in our experiments. It is worth noting that the semantic news encoder in our framework is \textit{plug-and-play}, which can be easily replaced by any other textual encoders or pretrained language models, e.g., BERT~\citep{BERT} or DeBERTa~\citep{Deberta}.

\subsection{News Graph Encoding Channel}\label{news_graph_encoding_channel}
In this section, we will explain the news semantic-augmented graph~(SAG) construction and graph context learning. Our motivation is to retrieve semantic-relevant news from training corpus and construct a semantic-augmented graph to enrich the semantics of the original single candidate news.

\subsubsection{News Graph Construction}\label{news_graph_construction}
\textbf{Semantic-relevant News Retrieval.} Pretrained language models~(PLM) have achieved remarkable performance~\citep{SBERT, multilingual-SBERT} on semantic textual similarity~(STS) benchmarks. Motivated by \citet{RAG}, we utilize a PLM $\phi(\cdot)$ to retrieve semantic-relevant news from training news corpus\footnote{Specifically, we use pretrained mpnet-base-v2~\citep{mpnet} in the Sentence Transformers library \url{https://www.sbert.net/docs/pretrained_models.html} as the news retrieval PLM.} to augment the semantic information of the original single candidate news. In the retrieval process, the semantic similarity score $s_{i,j}$ of news $n_i$ and $n_j$~(corresponding texts $T_i$ and $T_j$) is computed by the similarity function $sim(\cdot,\cdot)$:
\begin{equation}\label{eq2}
s_{i,j}=sim\big(n_i,n_j\big)=\textrm{cosine}\big(\phi(T_i),\phi(T_j)\big)
\end{equation}
\textbf{Semantic-augmented Graph~(SAG).} For the original candidate news $n_{can}$, we initialize it as the root node $v_{0}$ of the semantic-augmented news graph $G_{n}$. We build $G_{n}$ by repeatedly extending semantic-relevant neighboring nodes to existing nodes of $G_{n}$. In each graph construction step, for an existing node $v_i$~(correspoding news $N_{i}$) of $G_{n}$, $M$ news documents $\{N_{j}\}^{M}_{j=1}$ are retrieved from the news corpus $\{N_{C}\}$ with the highest semantic similarity scores $\{s_{i,j}\}^{M}_{j=1}$. We extend the nodes $\{v_{j}\}^{M}_{j=1}$ as neighboring nodes to the node $v_i$ by adding bidirectional edge $\{e_{i,j}\}^{M}_{j=1}$ between them. To heuristically discover semantic-relevant news in higher-order relations, we repeatedly extend the semantic-relevant news nodes within $K$ hops from the root node. The scale of news graph $G_{n}$ is approximated to be $\mathcal{O}(M^K)$. Detailed SAG construction and qualitative analysis are provided in {\color{blue}\textbf{Appendix}}~\ref{sec:appendixA}.

\subsubsection{News Graph Context Extraction}\label{news_graph_context_extraction}
Given an SAG $G_{n}$ generated from the candidate news node $v_{0}$ with $N$ semantic-relevant news nodes $\{v_{i}\}^{N}_{i=1}$, we use the semantic news encoder~(introduced in Section~\ref{news_encoder}) to extract their semantic representations as $h_{n,0}\in{\mathbb{R}^{d}}$ and $\{h_{n,i}\}^{N}_{i=1}\in{\mathbb{R}^{N\times{d}}}$.

We aim to extract the graph context $c_{n}\in{\mathbb{R}^{d}}$ which augments the semantics of the candidate news $n_{can}$ by aggregating the information of $G_{n}$. We consider the original semantics of the candidate news preserved in the root node $v_{0}$ and regard the local graph context as $h^{L}_n=h_{n,0}\in{\mathbb{R}^{d}}$. Besides, we employ an attention module to aggregate the global graph context $h^{G}_{n}\in{\mathbb{R}^{d}}$ from the semantic-relevant news nodes to encode the overall semantic information of $G_{n}$. In the attention module, we regard the root node embedding $h_{n,0}$ as the query and the semantic-relevant news node embeddings $\{h_{n,i}\}^{N}_{i=1}$ as the key-value pairs:
\begin{equation*}
    e_{i} = \frac{\,(h_{n,0}\textbf{W}^{Q}_{n})(h_{n,i}\textbf{W}^{K}_{n})^{T}\,}{\sqrt{d\;}}
\end{equation*}
\vskip -3.75mm
\begin{equation*}
    {\mathlarger\alpha}_{i} = \mathrm{softmax}(e_{i})=\frac{\mathrm{exp}(e_{i})}{\,\sum^{N}_{j=1}\mathrm{exp}(e_{j})\,}
\end{equation*}
\vskip -2.75mm
\begin{equation}\label{eq3}
    h^{G}_{n} = \sum_{i=1}^{N}{\mathlarger\alpha}_{i}h_{n,i}
\end{equation}
, where $\textbf{W}^{Q}_{n}\in{\mathbb{R}^{d\times{d}}}$ and $\textbf{W}^{K}_{n}\in{\mathbb{R}^{d\times{d}}}$ are parameter matrices. We integrate the local and global graph contexts by a simple feed-forward gating network $\textrm{FFN}_{g}(\cdot)$ to derive the news graph context $c_{n}$:
\begin{equation}\label{eq4}
    c_{n} = \textrm{FFN}_{g}\bigl([h^{L}_{n};h^{G}_{n}]\bigr)
\end{equation}
The parameters of news graph context extractor are shared among different graph layers of \textit{DIGAT}~(the user graph context extractor in Section~\ref{user_graph_context_extraction} also shares parameters likewise).

\subsection{User Graph Encoding Channel}\label{user_graph_encoding_channel}
\subsubsection{User Graph Construction}\label{user_graph_construction}
Motivated by \citet{CNE-SUE} and \citet{HGP}, we model user history with graph structure to encode multi-levels of user interests. We build a user graph $G_{u}$ containing \textbf{news nodes} and \textbf{topic nodes}: ($1$) For a user's clicked news $H_{u}=[n_{1},n_{2},...,n_{|H|}]$, we treat it as a set of news nodes for news-level user interest representation. ($2$) For the clicked news $n_{j}$, it is pertaining to a specific news topic\footnote{For example, in the \textit{MIND} dataset~\citep{MIND}, each news has a topic category~(e.g., \textit{Sports} and \textit{Entertainment}).} $t(i)$. We treat the clicked news topics as topic nodes for topic-level user interest representation. To capture the interaction among news and topics, we introduce three types of edges:

\textbf{News-News Edge.} News nodes with the same topic category~(e.g., \textit{Sports}) are fully connected. In this way, we can capture the relatedness among clicked news with news-level interaction.

\textbf{News-Topic Edge.} We model the interaction between clicked news and topics by connecting news nodes to their pertaining topic nodes.

\textbf{Topic-Topic Edge.} Topic nodes are fully connected. In this way, we can capture the overall user interests with topic-level interaction.

\subsubsection{User Graph Context Extraction}\label{user_graph_context_extraction}
Given the user history $H_{u}=[n_{1},n_{2},...,n_{|H|}]$, we employ the semantic news encoder~(introduced in Section~\ref{news_encoder}) to learn the historical news embeddings $h^{n}_{u}=[h^{n}_{u,1},h^{n}_{u,2},...,h^{n}_{u,|H|}]\in\mathbb{R}^{|H|\times{d}}$. Given $|t(\cdot)|$ topics indicated by the clicked news, the topic nodes are embedded into learnable embeddings $h^{t}_{u}=[h^{t}_{u,1},h^{t}_{u,2},...,h^{t}_{u,|t(\cdot)|}]\in\mathbb{R}^{|t(\cdot)|\times{d}}$. The user node embeddings are $h_{u}=[h^{n}_{u},h^{t}_{u}]$.

Motivated by \citet{HieRec}, we extract the graph context $c_{u}\in{\mathbb{R}^{d}}$ in a hierarchical way. First, we employ an attention module to learn the topic representation $\tilde{h}_{t(i)}\in\mathbb{R}^{d}$ of the topic $t(i)$. The topic-attention module regards the news graph context $c_{n}$ as the query and the news embeddings $\{h^{n}_{u,j}\}_{n_{j}\in{t(i)}}$ of topic $t(i)$ as the key-value pairs:
\begin{equation}\label{eq5}
    \tilde{h}_{t(i)} = \mathrm{Attn}\big(\mathrm{Q}\scriptstyle{=}\textstyle c_{n}, \mathrm{K}\scriptstyle{=}\textstyle\{h^{n}_{u,j}\},\mathrm{V}\scriptstyle{=}\textstyle\{h^{n}_{u,j}\}\big)
\end{equation}
Then, we employ another attention module to extract the user graph context $c_{u}\in\mathbb{R}^{d}$. The user-attention module regards the news graph context $c_{n}$ as the query and the learned topic representations $\{\tilde{h}_{t(i)}\}^{|t(\cdot)|}_{i=1}$ as the key-value pairs:
\begin{equation}\label{eq6}
    c_{u} = \mathrm{Attn}\big(\mathrm{Q}\scriptstyle{=}\textstyle c_{n}, \mathrm{K}\scriptstyle{=}\textstyle\{\tilde{h}_{t(i)}\}, \mathrm{V}\scriptstyle{=}\textstyle\{\tilde{h}_{t(i)}\}\big)
\end{equation}
$\mathrm{Attn}(\mathrm{Q},\mathrm{K},\mathrm{V})$ in Eq. (\ref{eq5}) and (\ref{eq6}) denotes the standard attention module with Query/Key/Value. We implement $\mathrm{Attn}(\mathrm{Q},\mathrm{K},\mathrm{V})$ as scaled dot-product attention~\citep{transformer} in our experiments.

\input{figs/model}
\subsection{Dual-Graph Interaction}\label{dual_graph_interacion}
In news graph $G_{n}$, node embeddings $\{h_{n,i}\}^{|G_{n}|}_{i=0}$ contain the information of augmented candidate news semantics. In user graph $G_{u}$, node embeddings $\{h_{u,i}\}^{|G_{u}|}_{i=0}$ contain the information of user history. We learn informative news and user graph embeddings by aggregating neighboring node information with stacked graph attention layers~\citep{GAT}. Most notably, our dual-graph interaction model aims at facilitating effective feature interaction between the news and user graphs. By effective dual-graph feature interaction, accurate news-user representation matching can be achieved. In the dual-graph interaction, the ($l+1$)-layer news node embeddings $h^{(l+1)}_{n}$ is updated based on the $l$-layer news node embeddings $h^{(l)}_{n}$ and user graph context $c^{(l)}_{u}$ jointly~(vice versa to update the user node embeddings $h^{(l+1)}_{u}$), as illustrated in Figure~\ref{fig:model}.

We illustrate the news node embedding update process for example. We first perform a linear trans-formation on the $l$-layer news node embedding $h^{(l)}_{n,i}$ to derive higher-level graph features $\hat{h}_{n,i}$:
\begin{equation}\label{eq7}
    \hat{h}_{n,i} = \hat{\textbf{W}}^{l}_{n}h^{(l)}_{n,i}+\hat{\textbf{b}}^{l}_{n},
\end{equation}
, where $\hat{\textbf{W}}^{l}_{n}\in{\mathbb{R}^{d\times{d}}}$ and $\hat{\textbf{b}}^{l}_{n}\in{\mathbb{R}^{d}}$ are learnable.

In order to learn news node embeddings interacting with user graph, we incorporate the user graph context $c^{(l)}_{u}$ into news graph attention computation. For news node $i$ and node $j\in\mathcal{N}^{n}_{i}$ (where $\mathcal{N}^{n}_{i}$ is the neighborhood of node $i$), we incorporate user graph context $c^{(l)}_{u}$ into computing the attention key vector $K_{i,j}$. We use a feed-forward network $\textrm{FFN}^{(l)}_{n}$ to compute $K_{i,j}$ based on the fused information of $c^{(l)}_{u}$, $h^{(l)}_{n,i}$ and $h^{(l)}_{n,j}$. The news graph attention coefficient $\alpha_{i,j}$ is computed aware of user graph context:
\vskip -1mm
\begin{equation}\label{eq8}
    K_{i,j} = \textrm{FFN}^{(l)}_{n}\Big(\big[c_{u}^{(l)};h^{(l)}_{n,i};h^{(l)}_{n,j}\big]\Big)
\end{equation}
\vskip -4mm
\begin{equation}\label{eq9}
    {\mathlarger\alpha}_{i,j} = \frac{\mathrm{exp}\Big(\mathrm{LeakyReLU}\big(\textbf{a}_{n}^{T}K_{i,j}\big)\Big)}{\sum\limits_{k\in\mathcal{N}^{n}_{i}}\mathrm{exp}\Big(\mathrm{LeakyReLU}\big(\textbf{a}_{n}^{T}K_{i,k}\big)\Big)}
\end{equation}
, where $\textbf{a}^{T}_{n}$ is a learnable attention weight vector. Finally, we aggregate the neighboring node embeddings with attention coefficient $\alpha_{i,j}$, followed by ReLU activation. Residual connection is applied to mitigate gradient vanishing in deep graph layers:
\vskip -4mm
\begin{equation}\label{eq10}
    h^{(l+1)}_{n,i} = \mathrm{ReLU}\bigg(\sum_{j\in\mathcal{N}^{n}_{i}}{\mathlarger\alpha}_{i,j}\hat{h}_{n,j}\bigg) + h^{(l)}_{n,i}
\end{equation}
\vskip -2.5mm
The news and user graph contexts $c^{(l)}_{n}$ and $c^{(l)}_{u}$ are extracted from the $l$-layer graph node embeddings as described in Section~\ref{news_graph_context_extraction} and \ref{user_graph_context_extraction}. We summarize Eq.~(\ref{eq7}) to (\ref{eq10}) as the news node embedding update function $\Phi^{(l)}_{n}$:
\vskip -3mm
\begin{equation}\label{eq11}
    h^{(l+1)}_{n,i} = {\mathlarger{\mathlarger\Phi}}^{(l)}_{n}\Big(c^{(l)}_{u},h^{(l)}_{n,i},\big\{h^{(l)}_{n,j}\big\}_{j\in\mathcal{N}^{n}_{i}}\Big)
\end{equation}
Similarly, the update function of user node embeddings is formulated as $\Phi^{(l)}_{u}$:
\vskip -3mm
\begin{equation}\label{eq12}
    h^{(l+1)}_{u,i} = {\mathlarger{\mathlarger\Phi}}^{(l)}_{u}\Big(c^{(l)}_{n},h^{(l)}_{u,i},\big\{h^{(l)}_{u,j}\big\}_{j\in\mathcal{N}^{u}_{i}}\Big)
\end{equation}
The dual-graph interaction can be viewed as an iterative process that performs ($1$) user graph context-aware attention to update news node embeddings and ($2$) news graph context-aware attention to update user node embeddings. We model the dual interaction with $L$ stacked layers. The final layers of news and user graph contexts $c^{L}_n$ and $c^{L}_u$ are adopted as news and user graph representations $r_{n}$ and $r_{u}$ which refine the news and user graph information with deep feature interaction. Algorithm \ref{alg:dual_graph_interaction} illustrates the dual-graph interaction process.
\input{algorithms/dual_graph_interaction_single_column}

\subsection{Click Prediction and Model Training}
With the news and user graph representations $r_{n}$ and $r_{u}$, our model aims to predict the matching score $\hat{s}_{n,u}$ which signals how likely user $u$ will click news $n$. The matching score between news and user representations is simply computed by dot product as $\hat{s}_{n,u}=r_{n}^{T}r_{u}$.

Following \citet{NAML, NPA, NRMS}, we adopt negative sampling strategy to train our model. For the user behavior that user $u$ had clicked news $n_{i}$, we compute the click matching score as $\hat{s}^{+}_{i}$ for $n_{i}$ and $u$. Besides, we randomly sample $S$ non-clicked news $[n_{1},n_{2},...,n_{S}]$ from the user's behavior log and compute the negative matching scores as $[\hat{s}^{-}_{i,1},\hat{s}^{-}_{i,2},...,\hat{s}^{-}_{i,S}]$. We optimize the NCE loss $\mathcal{L}$ over the training dataset $\mathcal{D}$ in model training:
\begin{equation}\label{eq13}
    \mathcal{L}=-\sum_{i=1}^{|\mathcal{D}|}{log\frac{\mathrm{exp}(\hat{s}^+_i)}{\mathrm{exp}(\hat{s}^+_i)+\sum_{j=1}^{S}{\mathrm{exp}(\hat{s}^-_{i,j})}}}
\end{equation}

\input{tables/main_experiments}
\section{Experiments}
\subsection{Dataset and Experiment Settings}\label{experiment_dataset_settings}
We conduct experiments on the real-world benchmark dataset \textit{MIND}~\citep{MIND}. \textit{MIND} is constructed from anonymized user behavior logs of Microsoft News with two versions of \textit{MIND-large} and \textit{MIND-small}. \textit{MIND-large} contains $1$ million anonymized users with user-click impression logs of $6$ weeks from October $12$ to November $22$, $2019$. The training and dev sets contain the impression logs of the first $5$ weeks, and the last week's impression logs are reserved for test. \textit{MIND-small} consists of $50000$ users, which are randomly sampled from \textit{MIND-large} with the impression logs.

Following previous works~\citep{FIM, HieRec}, we use news titles with the maximum length of $32$ words for news textual encoding. The user history includes $50$ news items they have recently clicked. The news word embeddings are $300$-dimensional and initialized from the pretrained Glove embeddings~\citep{glove}. Following~\citet{LSTUR}, we set the number of negative news samples $S$ to be $4$. For our model parameters, the news representation dimension $d$ is set as $400$ for fair comparison to baselines. The number of neighboring nodes $M$ and hops $K$ are $5$ and $2$, respectively. We set the number of dual-graph interaction layers as $L=3$. We use Adam optimizer~\citep{Adam} with the learning rate of $1$e-$4$ to train our model. Following~\citet{MIND}, we employ the recommendation ranking metrics AUC, MRR, nDCG@$5$, and nDCG@$10$ to evaluate model performance.

\subsection{Compared Methods}\label{compared_methods}
We compare our model with the state-of-the-art news recommendation methods: ($1$) \textit{GRU} \citep{DAE_RNN}, learning user representations from a sequence of clicked news with a GRU network; ($2$) \textit{DKN} \citep{DKN}, using a knowledge-aware CNN to learn news representations from both news texts and knowledge entities; ($3$) \textit{NPA} \citep{NPA}, encoding news and user representations with personalized attention networks; ($4$) \textit{NAML} \citep{NAML}, learning news representations from news titles, bodies, categories and subcategories with multi-view attention networks; ($5$) \textit{LSTUR} \citep{LSTUR}, jointly modeling long-term user embeddings and short-term user interests learned by a GRU network; ($6$) \textit{NRMS} \citep{NRMS}, encoding informative news and user representations with multihead self-attention networks; ($7$) \textit{FIM} \citep{FIM}, encoding news content with dilated convolutional networks and modeling user interest matching with $3$D convolutional networks; ($8$) \textit{HieRec} \citep{HieRec}, modeling user interests in a three-level hierarchy and performing multi-grained matching between candidate news and hierarchical user interest representations.

We also compare our model with competitive graph-based methods: ($9$) \textit{GERL} \citep{GERL}, modeling the news-user relatedness with a bipartite graph, which enhances news and user representations by aggregating neighboring node information; ($10$) \textit{GNewsRec} \citep{GNewsRec}, using graph neural networks~(GNN)~\citep{GNN} and attentive LSTMs to jointly model users' long-term and short-term interests; ($11$) \textit{User-as-Graph} \citep{HGP}, utilizing a heterogeneous graph pooling method to extract user representations from personalized heterogeneous behavior graphs.

\subsection{Main Experiment Results}\label{main_experiments}
Table~\ref{table:main_experiments} presents the main experiment results. We can observe that \textit{DIGAT} significantly outperforms previous SOTA methods~(i.e., methods \#$1$ to \#$8$) on the both datasets. This is because even though some baselines use topic categories or knowledge entities to enrich news information~(e.g., \textit{HieRec} learns news representations from both news texts and knowledge entities), the information entailed in single candidate news may be still insufficient. In contrast, \textit{DIGAT} can substantially enrich the semantic information of the single candidate news by SAG modeling, which provides more accurate signals of candidate news to match user interests. Besides, \textit{DIGAT} consistently outperforms three graph-based baselines. We find that \textit{GERL} is hard to model news-user interaction in test data, because most candidate news items in test data are fresh and have no click-interaction with users. Differently, \textit{DIGAT} models news and users with dual graph channels instead of a joint bipartite graph, which circumvents this \textit{cold-news} issue. Compared to \textit{GNewsRec} and \textit{User-as-Graph}, \textit{DIGAT} performs more effective feature interaction between the news and user graphs, which can enhance more accurate news-user representation matching.

\input{tables/SAG_effectiveness}
\subsection{Ablation Study on SAG Modeling}\label{SAG_modeling_ablation_study}
We examine the effectiveness of SAG modeling with three ablation experiments: ($1$) \underline{\textbf{w/o SA}}. To examine the effectiveness of semantic-augmentation (SA) strategy, we remove SAG from \textit{DIGAT} and learn single candidate news representation instead. ($2$) \underline{\textbf{TF-IDF SA}}. To inspect the function of the news retrieval PLM $\phi(\cdot)$ in SAG construction (see Section~\ref{news_graph_construction}), we replace $\phi(\cdot)$ with a TF-IDF syntactic feature extractor to retrieve relevant news. ($3$) \underline{\textbf{Seq SA}}. To examine the effectiveness of graph-based SA, we conduct controlled experiments by arranging the semantic-relevant news in a sequential form and extracting the news sequence context similar to Eq.~(\ref{eq3}) and (\ref{eq4}). Experiments in this section and the following sections are on \textit{MIND-small}.

Table~\ref{table:SAG_effectiveness} shows the experiment results. We can see that abandoning the SA strategy~(\textbf{w/o SA}) leads to the largest performance drop, as \textbf{TF-IDF SA} and \textbf{Seq SA} also yield better performance than \textbf{w/o~SA}. This validates the effectiveness of SA strategy to enrich candidate news semantics and further enhance news recommendation. \textbf{TF-IDF SA} underperforms \textit{DIGAT} by a considerable margin. We infer that the TF-IDF features can only measure news similarity at the syntactic level, which may not be able to accurately retrieve semantic-relevant news for SAG construction. In contrast, PLM can accurately measure news similarity at the semantic level and help retrieve more relevant news to enhance SAG modeling. It reveals that accurately retrieving semantically relevant news is the key to candidate news semantic-augmentation. Besides, \textbf{Seq SA} is suboptimal compared to the original graph-based SA. This is because the graph-based SA method can accurately model the relatedness among the candidate news and semantic-relevant news with multi-neighbor and multi-hop graph structure, which further improves the effectiveness of the SA strategy.

\input{figs/dual-graph-ablation-study}
\input{figs/SAG-settings}
\subsection{Ablation Study on Graph Interaction}\label{dual_graph_ablation_study}
To examine the effectiveness of dual-graph interaction, we design the following ablation experiments: ($1$) \underline{\textbf{w/o Interaction}}. We employ the vanilla graph attention networks (GAT) \citep{GAT} to learn news and user graph embeddings, respectively, without interaction between dual graphs. ($2$) \underline{\textbf{News Graph w/o Inter}}. The news graph embedding update layers are replaced with vanilla GAT layers. Concretely, Eq. (\ref{eq11}) is modified into $h^{(l+1)}_{n,i} = {\bar\Phi}^{(l)}_{n}\big(h^{(l)}_{n,i},\{h^{(l)}_{n,j}\}_{j\in\mathcal{N}^{n}_{i}}\big)$, where ${\bar\Phi}^{(l)}_{n}$ is the standard GAT graph embedding update function without feature interaction with user graph context. ($3$) \underline{\textbf{User Graph w/o Inter}}. Similar to ($2$), we replace the user graph embedding update layers with vanilla GAT layers.

Figure~\ref{fig:dual_graph_ablation} shows the performance of the ablation models. We can see that \textbf{w/o Interaction} underperforms the other three models with graph interaction modeling. It indicates that feature interaction between candidate news and users is necessary to enhance news recommendation. We also observe that removing user graph interaction (\textbf{User Graph w/o Inter}) leads to more performance drop than \textbf{News Graph w/o Inter}, which implies that user graph interaction may contribute more to our model. Moreover, \textit{DIGAT} surpasses the two single graph interaction ablations by a significant margin, validating the effectiveness of modeling dual-graph feature interaction in an iterative manner.

\subsection{Analysis on SAG Parameters} \label{SAG_settings_analysis}
We investigate two key parameters of SAG, i.e., the number of node neighbors $M$ and hops $K$. Figure \ref{fig:SAG_settings} shows the effect of different $M$ and $K$ settings.

As shown in Figure \ref{fig:SAG_settings}(a), \textit{DIGAT} performance continues rising as $M$ increases from $1$ to $5$. This indicates that with more semantic-relevant news incorporated, SAG can leverage more sufficient semantic information to augment the candidate news representations. On the other hand, the model performance slightly declines as $M>5$. The reason could be twofold. First, as the scale of SAG grows larger, it becomes more challenging for the model to distill the global graph context of SAG~(see Section \ref{news_graph_context_extraction}). Second, as $M$ becomes too large, it is inevitable to retrieve more noisy news in the SAG construction process, which may adversely affect SAG modeling. From Figure~\ref{fig:SAG_settings}(b), we observe that $K=2$ is the optimal hop setting. This may be because two hops of SAG can heuristically capture more useful semantic-relevant news information than simple one-hop modeling, while higher-hop extension may introduce too much irrelevant news and interfere with accurate semantic augmentation for candidate news. In general, we select $M=5$ and $K=2$ for SAG construction\footnote{The SAG construction on \textit{MIND-large} can be preprocessed in $20$ minutes on Intel Xeon(R) Gold $6226$R CPU @ $2.90$ GHz with Nvidia RTX $3090$ GPU before model training.}.

\input{figs/dual-graph-depth}
\subsection{The Number of Dual-Graph Layers} \label{dual_graph_layers}
We study the effect of the number of dual-graph layers $L$ in \textit{DIGAT}. The results are presented in Figure~\ref{fig:dual_graph_depth}. We can see that the model performance first keeps rising when $L$ increases from $1$ to $3$. It suggests that deep feature interaction between news and user graphs is useful to improve recommendation performance, as it can model the news and user representation matching process in a more fine-grained way. We also observe that further increasing $L$ hurts the model performance. It may be caused by the unstable gradient in training the deep dual-graph architecture, as we empirically find that gradient clipping~\citep{gradient_clipping} is indispensable to avoid loss diverging in \textit{DIGAT} training, in cases when the dual-graph layers become too deep~(i.e., $L\geq6$).

\section{Conclusion}
In this work, we present a dual-graph interaction framework for news recommendation. In our approach, a graph enhanced semantic-augmentation strategy is employed to enrich the semantic information of candidate news. Moreover, we design a dual-graph interaction mechanism to achieve effective feature interaction between news and user graphs, facilitating more accurate news and user representation matching. Our approach advances the state-of-the-art news recommendation methods on the benchmark dataset \textit{MIND}. Extensive experiments and further analyses validate that SAG modeling and dual-graph interaction can effectively improve news recommendation performance.

\section{Limitations}
In this section, we discuss the limitations of our approach. First, since \textit{DIGAT} models dual-interaction between news and user graph features iteratively, the inference efficiency is a concern. We compare the model size and inference run-time of experimental methods in Table~\ref{table:efficiency}. The news representations (see Section~\ref{news_encoder}) of all methods, except \textit{NPA}\footnote{This is because the personalized user ID embeddings are premised in computing \textit{NPA} news representations.}, are pre-computed and cached for fast inference. As \textit{DIGAT} is scalable with the dual-graph depth $L$, we also evaluate \textit{DIGAT} on $L=1$ and $2$.

In terms of model size, \textit{DIGAT} is larger than the first eight models in Table~\ref{table:efficiency}. Compared to \textit{DIGAT} ($L=1,2$), we can see that the parameter growth comes from the stacked graph layers. We also find that embedding layers contain considerable parameters\footnote{For example, \textit{GERL} needs additional embedding matrices to embed news and user IDs, \textit{NAML} needs larger word embedding matrices to encode news titles and bodies together.}, while \textit{DIGAT} does not need additional news and user ID embedding layers. In terms of inference time, \textit{DIGAT} runs slower than other models. We find that the computational overhead mostly comes from the iterative graph embedding update process in Eq.~(\ref{eq11}) and (\ref{eq12}). Nonetheless, this efficiency issue can be alleviated. Since \textit{DIGAT} is scalable with the dual-graph depth $L$, the trade-off between recommendation accuracy and efficiency can be made. We can scale down the dual-graph layers $L$ to reduce the model size and inference time with compromising performance. As shown in Table~\ref{table:efficiency}, when the dual-graph layers turn down to $L=1$, the performance of \textit{DIGAT} is also superior to baseline methods, while the parameter size and inference time are comparable to several baselines~(e.g., \textit{FIM} and \textit{GNewsRec}). In industrial deployment, this trade-off can depend on specific requirements of computational resources.

Second, our approach is evaluated on the off-line experimental dataset. For online recommender services, \textbf{searching and retrieving real-time relevant news} by event-driven news clustering models \citep{news_event_clustering} to construct SAG is a more promising option than the static retrieval method. To this end, we will explore applying our approach to online applications in future work.

\input{tables/efficiency}
\section*{Acknowledgements}
We appreciate constructive comments from some anonymous reviewers. The research described in this paper is supported by Innovation \& Technology Commission HKSAR, under ITF Project No. PRP/054/21FX.

\bibliography{custom}
\bibliographystyle{acl_natbib}
\clearpage
\appendix
\input{algorithms/SAG_construction}
\section{Semantic-augmented Graph Construction and Qualitative Analysis}\label{sec:appendixA}

Algorithm~\ref{alg:SAG_construction} illustrates the procedure of semantic-augmented graph~(SAG) construction. First of all, the SAG $G_{n}$ is initialized from the root node $v_{0}$ which represents the original candidate news $n_{0}$.

The graph construction is performed by repeatedly extending semantic-relevant neighboring news nodes to existing nodes in $G_{n}$. In the \textbf{graph extension process}~(line $3$ to $18$ in Algorithm~\ref{alg:SAG_construction}), for an existing node $v_{i}$~(corresponding to news $n_{i}$) in $G_{n}$, we retrieve $M$ news documents $\{n_{j}\}_{j=1}^{M}$ from the news corpus\footnote{We use news in the \textit{train/news.tsv} data file to construct the news corpus in \textit{MIND-small} experiments, and news in the \textit{train\&dev/news.tsv} data files to construct the news corpus in \textit{MIND-large} experiments. The news documents in the test set are not included in the news corpus.} $\{N_{C}\}$ with the $M$ highest similarity scores $\{s_{i,j}\}_{j=1}^{M}$. The similarity score $s_{i,j}$ of news $n_{i}$ and $n_{j}$ is evaluated by a PLM $\phi(\cdot)$ with Eq.~(\ref{eq2}). We treat the retrieved news $\{n_{j}\}_{j=1}^{M}$ as news nodes $\{v_{j}\}_{j=1}^{M}$. For each node $v_{j}$, we extend it to $G_{n}$ as a neighboring node of $v_{i}$ by adding bidirectional edge $e_{i,j}$ between $v_i$ and $v_j$. To heuristically explore higher-order semantic-relevant news, news nodes in SAG are extended from the root node $v_{0}$ within $K$ hops at most.

\textbf{SAG Example.} Figure~\ref{fig:sag_example} demonstrates an example of SAG for the candidate news $n_{0}$ ``\textit{Should the NFL be able to fine players for criticizing officiating}''. Interestingly, from Figure~\ref{fig:sag_example}(b), we can see that there are many similar news articles in SAG, which refer to the same specific news event or person~(i.e., ``\textit{NFL}'' and ``\textit{fine players}'') from 
different {\color{blue}\textit{narrative points of view}}\footnote{\href{https://en.wikipedia.org/wiki/Narration}{https://en.wikipedia.org/wiki/Narration}.}. These semantic-relevant news articles are finely retrieved with the help of PLM retriever, forming explicit multi-neighbor and multi-hop graph structure. With the representation power of SAG, \textit{DIGAT} can learn more accurate relatedness of the relevant news texts and substantially enrich the semantic information of the original candidate news $n_{0}$.

\textbf{News Clustering Phenomenon.} From the SAG example shown in Figure~\ref{fig:sag_example}(a), we can observe that there exist many cyclic subgraphs~(i.e., news clusters), revealing the news clustering phenomenon in semantic space. These cyclic graph structures depict the similar news clusters in real-world distributions, consistent with the previous research \citep{news_clustering, news_event_clustering}. This news semantic clustering phenomenon also inspires the motivation of our work. % By incorporating the semantically-clustered news information, SAG can capture more robust and informative representations of the candidate news.

\textbf{Broader Impact.} On online news platforms, the {\color{blue}\textit{Related News}} is usually displayed along with the original news to users. It is worth mentioning that such {\color{blue}\textit{Related News}} on news platforms is practically retrieved from the news database by retrieval models in industrial practice~(Algorithm~\ref{alg:SAG_construction} can be seen as such an analogous retrieval process). As an alternative, we can also use the \textit{off-the-shelf real-time} {\color{blue}\textit{Related News}} on online news platforms to construct SAG. Furthermore, the SAG modeling strategy is also applicable to other text-based recommendation~(e.g., \textit{Twitter Feed Recommendation}). We will explore this direction in future work.

\input{figs/SAG-example}
\section{Supplementary Experiments on Semantic-Augmentation Strategy}\label{sec:appendixB}
We conduct supplementary experiments to investigate whether the semantic-augmentation (SA) strategy can be generalized for news recommendation task. To exclude the influence of \textit{DIGAT} itself, we choose to reinforce the baseline \textit{NRMS}~\citep{NRMS} with SA strategy, named \textit{NRMS-SA}\footnote{We choose \textit{NRMS} because it is a simple yet representative baseline for news recommendation and does not involve news-user interaction modeling for controlled experiments.}.

For \textit{NRMS-SA}, we use the PLM news retriever to retrieve $10$ semantic-relevant news articles for each candidate news. The candidate news and semantic-relevant news are encoded by the \textit{NRMS} news encoder. We follow Eq.~(\ref{eq3}) and (\ref{eq4}) to derive the local news representation $h^{L}_{n}$ and global news representation $h^{G}_{n}$, and finally learn the augmented candidate news representation. The \textit{NRMS} user encoder remains unchanged.

Table~\ref{table:SA} shows the experiment results, which indicate that semantic-augmentation strategy can also be applied to other news recommendation models and achieve substantial performance improvement. Interestingly, we find that the improvement on \textit{MIND-large} is more significant than on \textit{MIND-small}, as \textit{NRMS-SA} is even on par with the previous SOTA baseline~(i.e., \textit{User-as-graph}). We infer that it may be because the \textit{MIND-large} news corpus is an order of magnitude larger than \textit{MIND-small}, and hence it contains more semantic-relevant news for SAG modeling. The experiment results also suggest that augmenting the semantic representation of single candidate news by \textbf{relevant news information sources} is a promising direction to improve news recommendation performance.
\input{tables/SA}
\end{document}

%% file: figs/news-recommendation-framework.tex
\begin{figure}[t]
\centering
\includegraphics[height=36.5mm]{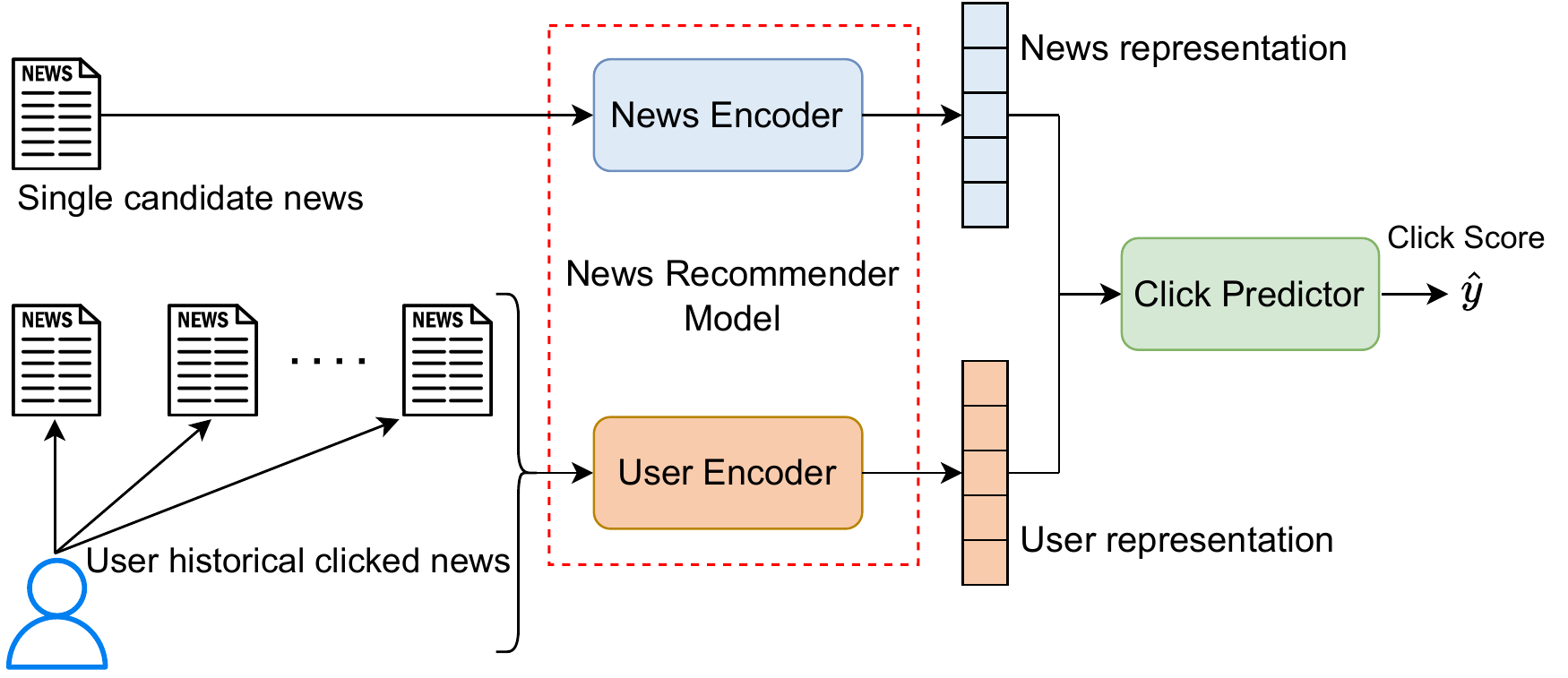}
\captionsetup{font=10pt}
\caption{
The typical news-user representation learning framework for news recommendation.
}
\vskip -1.225mm
\label{fig:news-recommendation-framework}
\end{figure}

%% file: figs/model.tex
\begin{figure}[t]
\centering
\includegraphics[height=60mm,width=84mm]{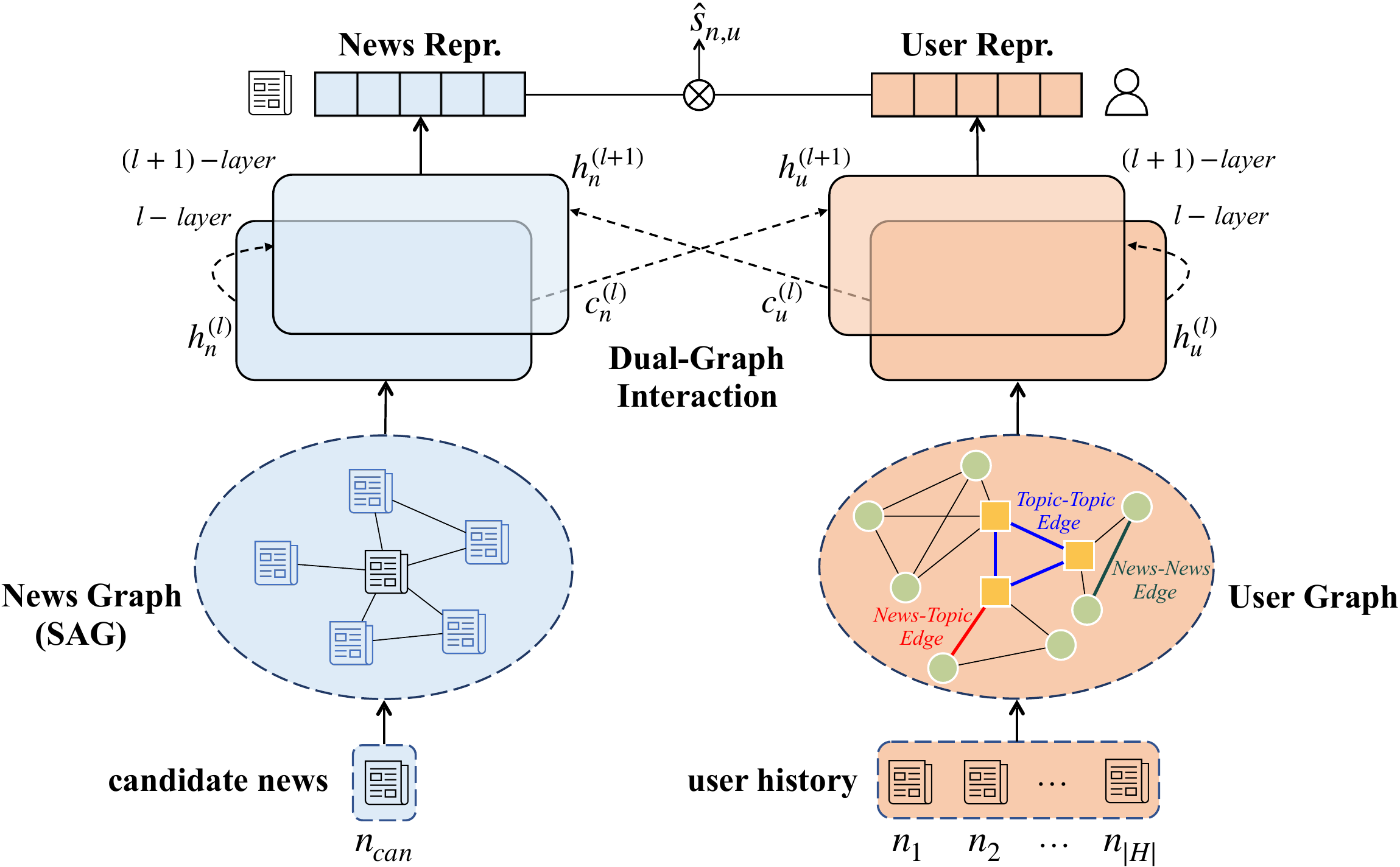}
\captionsetup{font=10pt}
\caption{
The overall architecture of DIGAT framework.
}
\label{fig:model}
\end{figure}

%% file: algorithms/dual_graph_interaction_single_column.tex
\begin{algorithm}[t]
\caption{News-User Graph Interaction} \label{alg:dual_graph_interaction}
\begin{algorithmic}[1]
\algsetup{linenosize=\small}
\small

\REQUIRE news node embeddings $h^{0}_{n}=\{h^{0}_{n,i}\}_{i=0}^{|G^n|}$, \\
         $\quad$ user $\;\!$ node embeddings $h^{0}_{u}=\{h^{0}_{u,i}\}_{i=0}^{|G^u|}$, \\
         $\quad$ number of dual-graph layers $L$.

\ENSURE news graph representation $r_{n}$, \\
        $\quad \;\;\,$ user $\;\!$ graph representation $r_{u}$.
\STATE Initialize $c^{0}_n$ from $h^{0}_n$ with Eq.~(\ref{eq3}) - (\ref{eq4}).
\STATE Initialize $c^{0}_u$ from $h^{0}_u\:\!$ with Eq.~(\ref{eq5}) - (\ref{eq6}).

\FOR{$l=0,1,...,L-1$}
\STATE Compute the ($l+1$)-layer news node embeddings $h^{(l+1)}_{n}$ with Eq.~(\ref{eq11}).
\STATE Compute the ($l+1$)-layer user $\:\!$ node embeddings $h^{(l+1)}_{u}$ with Eq.~(\ref{eq12}).
\STATE Compute the ($l+1$)-layer news graph context $c^{(l+1)}_n$ with Eq.~(\ref{eq3}) - (\ref{eq4}).
\STATE Compute the ($l+1$)-layer user $\:\!$ graph context $c^{(l+1)}_u$ with Eq.~(\ref{eq5}) - (\ref{eq6}).
\ENDFOR

\STATE $r_n = c^{L}_{n}$ and $r_u = c^{L}_{u}$.
\RETURN $r_n$, $r_u$
\end{algorithmic}
\end{algorithm}

%% file: tables/main_experiments.tex
\begin{table*}[t]
\centering
\resizebox{\textwidth}{!}{
\fontsize{9}{11}\selectfont
\begin{tabular}{l|l|cccc|cccc}
\hline
& & \multicolumn{4}{c|}{\textit{MIND-small}} & \multicolumn{4}{c}{\textit{MIND-large}} \\ \hline
\#& Method & AUC & MRR & nDCG@5 & nDCG@10 & AUC & MRR & nDCG@5 & nDCG@10 \\ \hline
1 & GRU & 61.51 & 27.46 & 30.11 & 36.61 & 65.42 & 31.24 & 33.76 & 39.47 \\
2 & DKN & 62.90 & 28.37 & 30.99 & 37.41 & 64.07 & 30.42 & 32.92 & 38.66 \\
3 & NPA & 64.65 & 30.01 & 33.14 & 39.47 & 65.92 & 32.07 & 34.72 & 40.37 \\
4 & NAML & 66.12 & 31.53 & 34.88 & 41.09 & 66.46 & 32.75 & 35.66 & 41.40 \\
5 & LSTUR & 65.87 & 30.78 & 33.95 & 40.15 & 67.08 & 32.36 & 35.15 & 40.93 \\
6 & NRMS & 65.63 & 30.96 & 34.13 & 40.52 & 67.66 & 33.25 & 36.28 & 41.98 \\
7 & FIM & 65.34 & 30.64 & 33.61 & 40.16 & 67.87 & 33.46 & 36.53 & 42.21 \\
8 & HieRec  & 67.83 & 32.78 & 36.31 & 42.49 & 69.03 & 33.89 & 37.08 & 43.01 \\
\hline
9 & GERL & 65.27 & 30.10 & 32.93 & 39.48 & 68.10 & 33.41 & 36.34 & 42.03 \\
10 & GNewsRec & 65.54 & 30.27 & 33.29 & 39.80 & 68.15 & 33.45 & 36.43 & 42.10 \\
11 & User-as-Graph$^\dagger$ & \textendash & \textendash & \textendash & \textendash & 69.23 & 34.14 & 37.21 & 43.04 \\
\hline
& DIGAT & 68.77 & 33.46 & 37.14 & 43.39 & 70.08	& 35.20	& 38.46	& 44.15 \\
\hline
\end{tabular}
}
\vskip -0.0975mm
\caption{Evaluation performance of all methods. Experiments of baseline \#$1$ to \#$10$ and \textit{DIGAT} are conducted $10$ times on \textit{MIND-small} and $5$ times on \textit{MIND-large}, respectively. We report the average performance. $^\dagger$Results of \textit{User-as-Graph} are directly copied from the previous work~\citep{HGP}. The performance improvements of \textit{DIGAT} compared to all baselines are significant (validated by Student's t-test with $p$-value $<0.01$).}
\label{table:main_experiments}
\end{table*}

%% file: tables/SAG_effectiveness.tex
\begin{table}[t]
\centering
\fontsize{9}{11}\selectfont
\begin{tabular}{l|c|c|c|c}
\hline
& AUC & MRR & \fontsize{8.875}{11}\selectfont{nDCG@5} & \fontsize{8.875}{11}\selectfont{nDCG@10} \\ \hline
w/o SA & 67.44 & 32.44 & 35.79 & 42.13 \\
TF-IDF SA & 67.82 & 32.65 & 36.25 & 42.49 \\
Seq SA & 68.29 & 33.01 & 36.60 & 42.91 \\
\hline
DIGAT & 68.77 & 33.46 & 37.14 & 43.39 \\
\hline
\end{tabular}
\caption{Experiment results of SAG modeling variants.}
\label{table:SAG_effectiveness}
\end{table}

%% file: figs/dual-graph-ablation-study.tex
\begin{figure}[t]
\centering
\includegraphics[width=77.25mm, height=52.75mm]{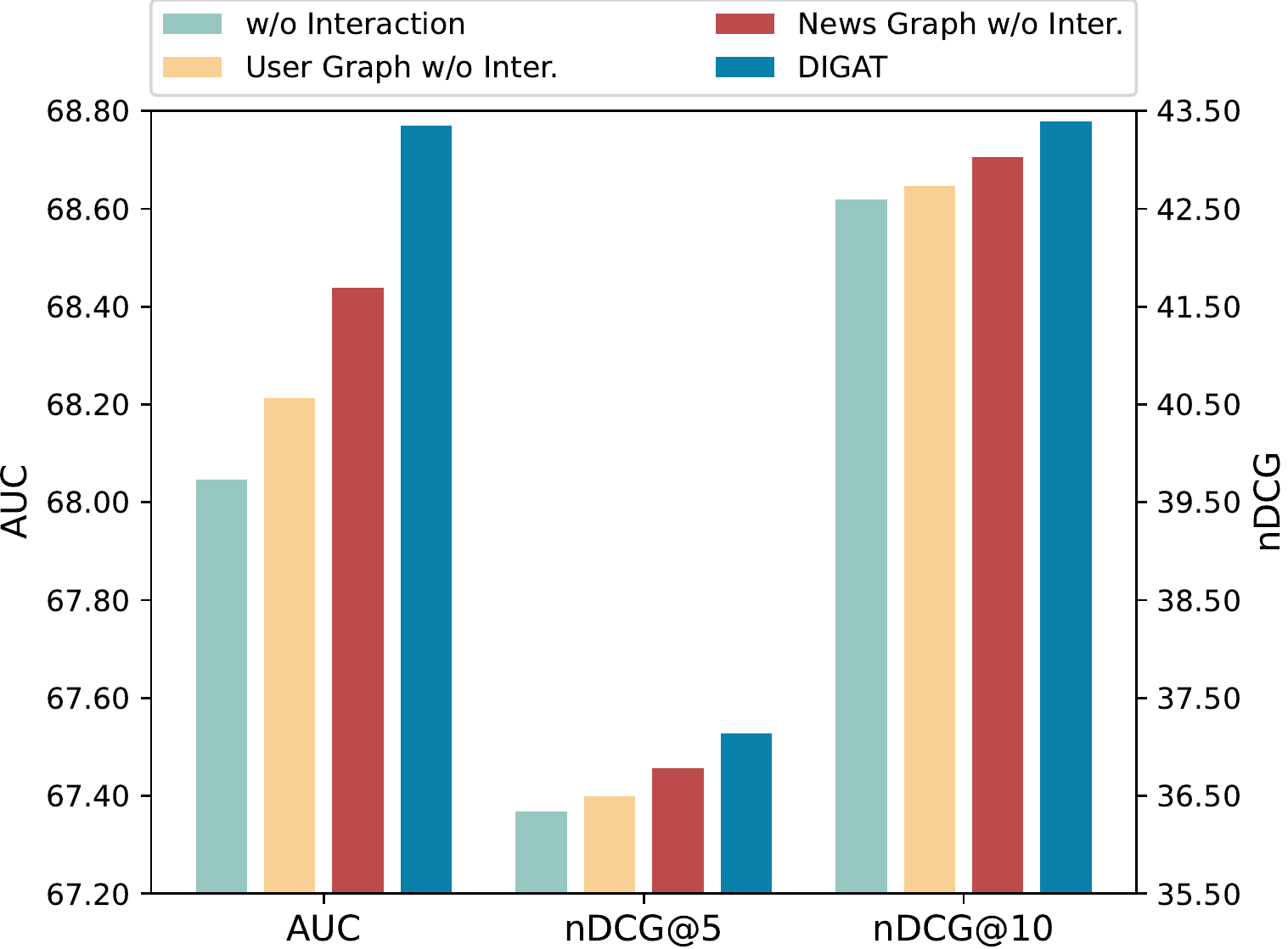}
\captionsetup{font=10pt}
\caption{
Ablation results on dual-graph interaction.
}
\label{fig:dual_graph_ablation}
\end{figure}

%% file: figs/SAG-settings.tex
\begin{figure*}[t]
\centering
\includegraphics[height=45mm, width=155mm]{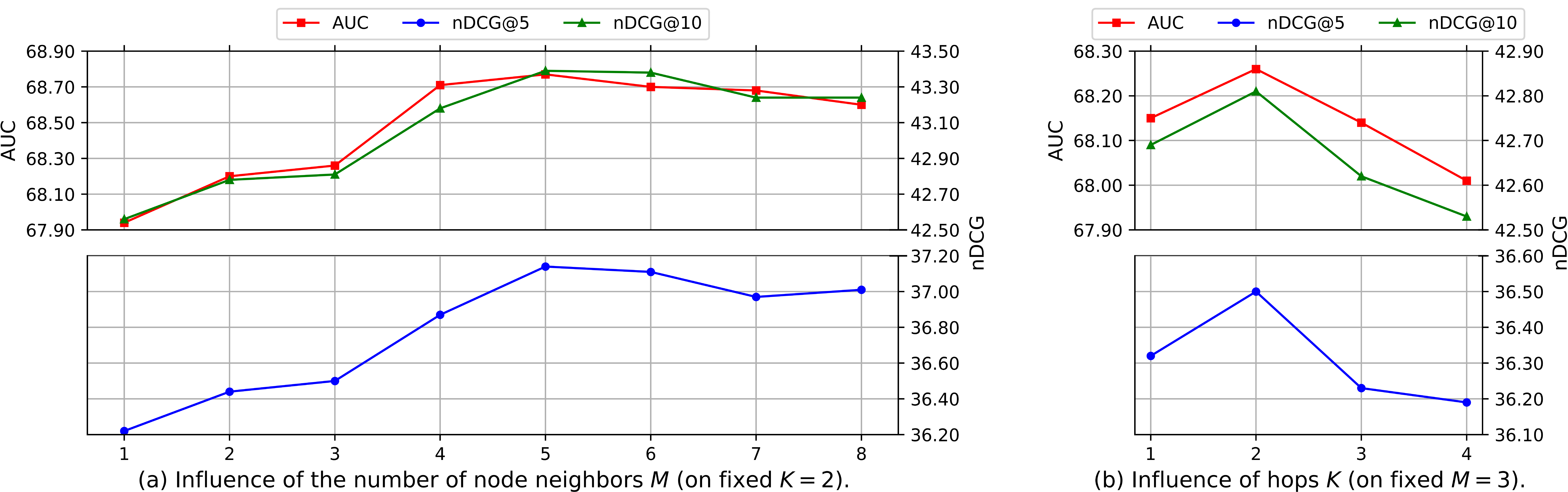}
\captionsetup{font=10pt}
\caption{
\textit{DIGAT} performance with different $M$ and $K$ settings of SAG.
}
\label{fig:SAG_settings}
\end{figure*}

%% file: figs/dual-graph-depth.tex
\begin{figure}[t]
\centering
\includegraphics[height=45mm]{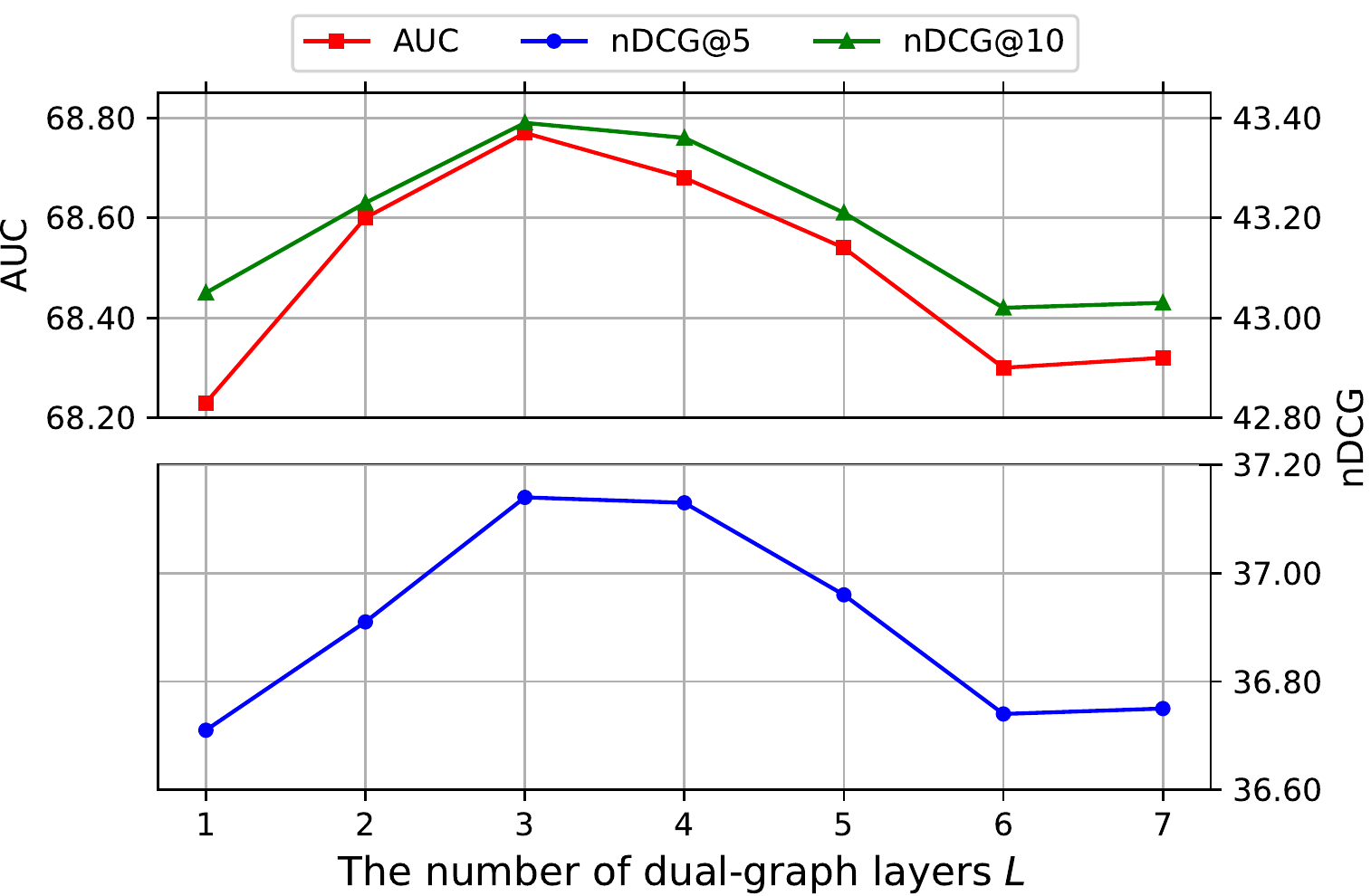}
\captionsetup{font=10pt}
\caption{
\textit{DIGAT} performance with different numbers of dual-graph layers $L$.
}
\label{fig:dual_graph_depth}
\end{figure}

%% file: tables/efficiency.tex
\begin{table}[t]
\centering
\fontsize{9}{11}\selectfont{
\begin{tabular}{l|c|c|c}
\hline
Method & Param.~(MB) & Run-time~(s) & AUC \\ \hline
GRU & 19.9 & 92.0 & 61.51\\ % 82.0
DKN & 23.3 & 80.5 & 62.90\\
NPA & 30.1 & 376.6 & 64.65\\
NAML & 39.3 & 80.1 & 66.12\\
LSTUR & 35.3 & 98.9 & 65.87\\ % 89.5
NRMS & 22.2 & 89.5 & 65.63\\ % 98.9
FIM & 22.5 & 495.7 & 65.34\\
HieRec & 31.7 & 106.9 & 67.83\\
\hline
GERL & 75.1 & 129.1 & 65.27\\
% 151.3
GNewsRec & 50.6 & 186.8 & 65.54\\
\hline
DIGAT & 40.3 & 598.5 & 68.77\\ % Unfortunately, we have to minus 10. After submission, we will try to remove the affine function user graph context extraction, which seems unnecessary.
\hline
DIGAT ($L=2$) & 35.4 & 426.3 & 68.60\\ % Unfortunately, we have to minus 10. After submission, we will try to remove the affine function user graph context extraction, which seems unnecessary.
DIGAT ($L=1$) & 30.5 & 250.7 & 68.23\\ % Unfortunately, we have to minus 10. After submission, we will try to remove the affine function user graph context extraction, which seems unnecessary.
\hline
\end{tabular}
}
\caption{Comparison of experimental methods' parameters and inference run-time. The \textbf{Run-time} column denotes the inference time on \textit{MIND-small} test set, which is averaged by $10$ times. All models are tested with the same batch size on Nvidia RTX $3090$.}
\label{table:efficiency}
\end{table}

%% file: algorithms/SAG_construction.tex
\begin{algorithm}[H]
    \caption{SAG Construction Procedure} \label{alg:SAG_construction}
    \begin{algorithmic}[1]
    \algsetup{linenosize=\small}
    \small
    % Input
    \REQUIRE candidate news $n_{0}$, news corpus $\{N_{C}\}$, news node\\
             $\quad$ neighbors $M$ and hops $K$.
    % Output
    \ENSURE semantic-augmented graph $G_{n}$. \\
    
    \STATE Regard $n_{0}$ as the root node $v_{0}$ of SAG. \\
    \STATE Initialize the node set $V\leftarrow\{v_{0}\}$ and edge set $E\leftarrow\{\}$. Define parent node set $P\leftarrow\{v_{0}\}$ and node-hop counter $\mathrm{hop}[v_{0}]=0$.

    \tcp{Graph extension process}
    \WHILE{$P\neq\emptyset$}
    \STATE Pop a node $v_{i}$ from $P$, then $P=P\setminus\{v_i\}$
    \STATE Retrieve $M$ news $\{n_{j}\}_{j=1}^{M}$ from the news corpus $\{N_{C}\}$ with the $M$ highest semantic similarity scores $\{s_{i,j}\}_{j=1}^{M}$ as nodes $\{v_{j}\}_{j=1}^{M}$ \\
    \FOR{$j=1,2,...,M$}
    \IF{$v_j \not\in V$}
    \STATE $V=V\cup\{v_j\}$
    \STATE $\mathrm{hop}[v_j]=\mathrm{hop}[v_i]+1$
    \IF{$\mathrm{hop}[v_j] < {K}$}
    \STATE $P=P\cup\{v_j\}$
    \ENDIF
    \ENDIF
    \IF{edge $e_{i,j} \not\in E$ }
    \STATE $E=E\cup\{e_{i,j}\}$
    \ENDIF
    \ENDFOR
    \ENDWHILE
    \STATE $G_{n}=\{V,E\}$.
    \RETURN $G_{n}$
    \end{algorithmic}
\end{algorithm}

%% file: figs/SAG-example.tex
\begin{figure*}[ht]
\centering
\includegraphics[height=100mm,width=160mm]{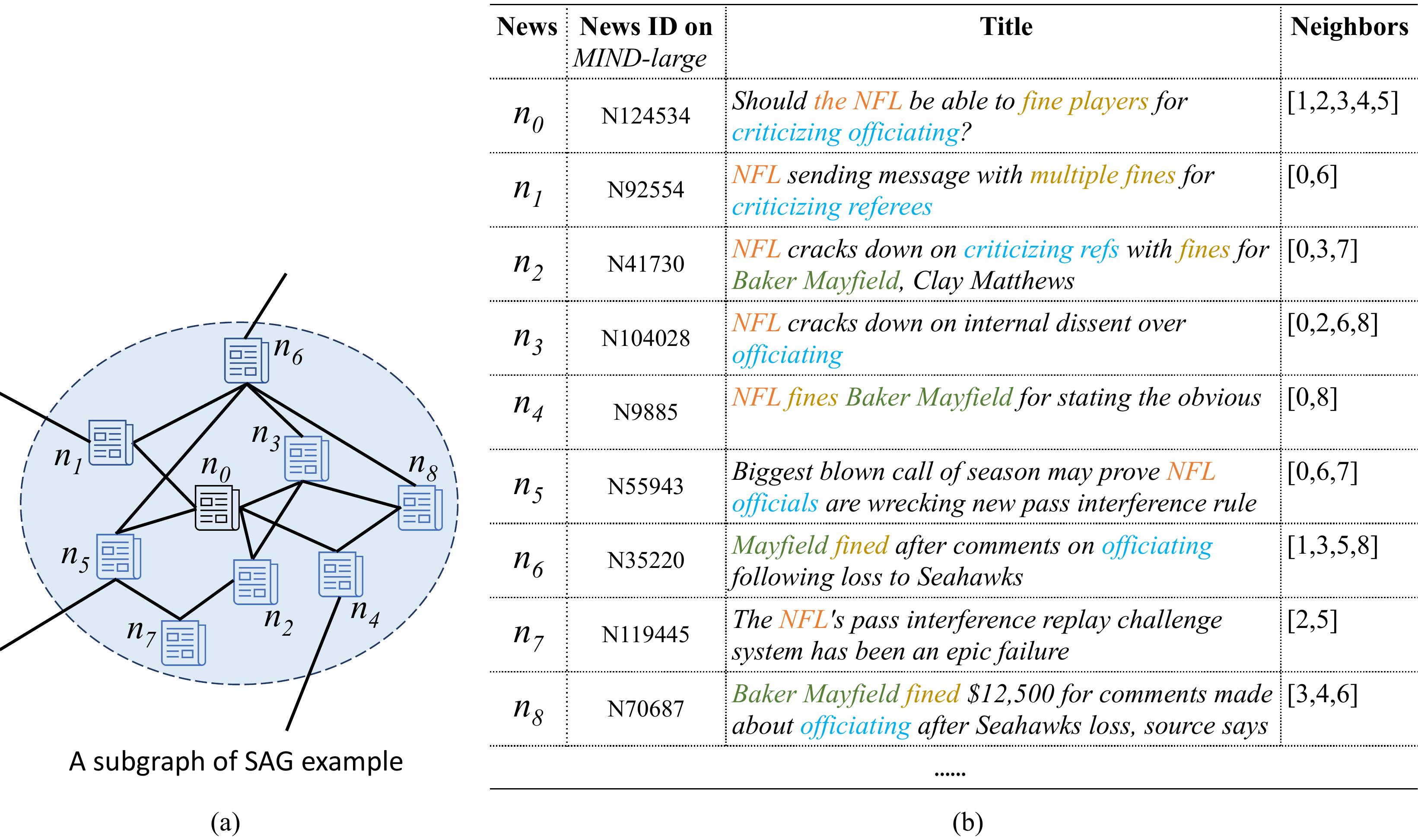}
\captionsetup{font=10pt}
\caption{
An example of SAG~($M=5$ and $K=2$) constructed from news $n_{0}$ on \textit{MIND-large} (news ID: N$124534$): (a) A subgraph of the example SAG including root node $n_0$ and semantic-relevant news node $n_i$ ($i=1,2,...,8$); (b) News in SAG and the corresponding title texts. For brevity, we only present an SAG subgraph of nodes and edges.
}
\label{fig:sag_example}
\end{figure*}

%% file: tables/SA.tex
\begin{table}[t]
\centering
\fontsize{9}{11}\selectfont
\begin{tabular}{l|cccc}
\hline
\multicolumn{5}{c}{\textit{MIND-small}} \\ \hline
Method & AUC & MRR & nDCG@5 & nDCG@10 \\ \hline
NRMS & 65.63 & 30.96 & 34.13 & 40.52 \\
NRMS-SA & 67.27	& 32.37 & 35.84	& 42.13 \\
HieRec  & 67.83 & 32.78 & 36.31 & 42.49 \\
\hline
\multicolumn{5}{c}{\textit{MIND-large}} \\ \hline
Method & AUC & MRR & nDCG@5 & nDCG@10 \\ \hline
NRMS & 67.66 & 33.25 & 36.28 & 41.98 \\
NRMS-SA & 69.31 & 34.39 & 37.56 & 43.27 \\
HieRec & 69.03 & 33.89 & 37.08 & 43.01 \\
User-as-Graph & 69.23 & 34.14 & 37.21 & 43.04 \\
\hline
\end{tabular}
\caption{Supplementary experiments on \textit{NRMS} reinforced with semantic-augmentation~(SA) strategy.}
\label{table:SA}
\end{table}